\documentclass[11pt,a4paper]{article}
\usepackage[hyperref]{emnlp2020}
\usepackage{times}
\usepackage{latexsym}

\usepackage{multirow}
\usepackage{longtable}
\usepackage{tabularx}
\usepackage{makecell}
\usepackage{graphicx}
\usepackage{caption}
\usepackage{subcaption}
\usepackage{enumitem}
\usepackage{amsmath}

\usepackage{booktabs}
\usepackage{chngpage}

\graphicspath{ {./figures/} }

\usepackage{todonotes}  
\usepackage[normalem]{ulem}
\newif\ifcomments
\commentstrue

\ifcomments
    \newcommand{\deepak}[1]{\textcolor{blue}{[DPK: #1]}}
    \newcommand{\ye}[1]{\textcolor{olive}{[Yanai: #1]}}
    \newcommand{\xikun}[1]{\textcolor{red}{[Xikun: #1]}}
    
    \definecolor{iansnotecolor}{RGB}{255, 218, 181}
    \newcommand{\iansidenote}[1]{
        \todo[color=iansnotecolor, size=\footnotesize]{%
        [\textbf{Ian:}] #1}%
    }
    \definecolor{iancolor}{RGB}{200, 100, 0}
    \newcommand{\ian}[1]{\textcolor{iancolor}{[Ian: #1]}}
    
    \newcommand{\maybedelete}[1]{\textcolor{red}{\sout{#1}}}
    
\else
    \newcommand*{\deepak}[1]{}
    \newcommand*{\ye}[1]{}
    \newcommand*{\xikun}[1]{}
    \newcommand*{\ian}[1]{}
    \newcommand*{\iansidenote}[1]{}

    \newcommand*{\maybedelete}[1]{}
    
\fi

\usepackage{microtype}

\aclfinalcopy 

\title{Do Language Embeddings Capture Scales?}

\author{Xikun Zhang\thanks{\hspace{1 mm} Both authors contributed equally.} \thanks{\hspace{1 mm} Work done during an internship at Google Research.}\\
  Stanford University \\
  \texttt{xikunz2@cs.stanford.edu} \\\And
  Deepak Ramachandran\footnotemark[1]\\
  Google Research \\
  \texttt{ramachandrand@google.com} \\\AND
  Ian Tenney\\
  Google Research \\
  \texttt{iftenney@google.com} \\\And
  Yanai Elazar\\
  Bar Ilan University, AI2 \\
  \texttt{yanaiela@gmail.com} \\\And
  Dan Roth\\
  University of Pennsylvania \\
  \texttt{danroth@seas.upenn.edu} \\}

\date{}

\begin{document}
\maketitle
\begin{abstract}
Pretrained Language Models (LMs) have been shown to possess significant linguistic, common sense and factual knowledge. One form of  knowledge that has not been studied yet in this context is information about the scalar magnitudes of objects. We show that pretrained language models capture a significant amount of this information but are short of the capability required for general common-sense reasoning. We identify contextual information in pre-training and numeracy as two key factors affecting their performance, and show that a simple method of canonicalizing numbers can have a significant effect on the results. \footnote{Code and models are available at \url{https://github.com/google-research-datasets/numbert}.}

\end{abstract}

\section{Introduction} \label{sec:intro}

The success of contextualized pretrained Language Models like BERT \citep{devlin2018bert} and ELMo \citep{peters2018deep}
on tasks like Question Answering and Natural Language Inference, has led to speculation that they are good at Common Sense Reasoning (CSR). 

On one hand, recent work has approached this question by measuring the ability of LMs to answer questions about physical common sense \cite{bisk2020} (``How to separate egg whites from yolks?"), temporal reasoning \cite{zhao2020} (``How long does a basketball game take?"), and numerical common sense \cite{lin2020birds}. On the other hand, after realizing some high-level reasoning skills like this may be difficult to learn from a language-modeling objective only, \cite{geva2020injecting} injects numerical reasoning skills into LMs by additional pretraining on automatically generated data. All of these skills are prerequisites for CSR.

\begin{figure}[!bt]
    \includegraphics[width=\columnwidth]{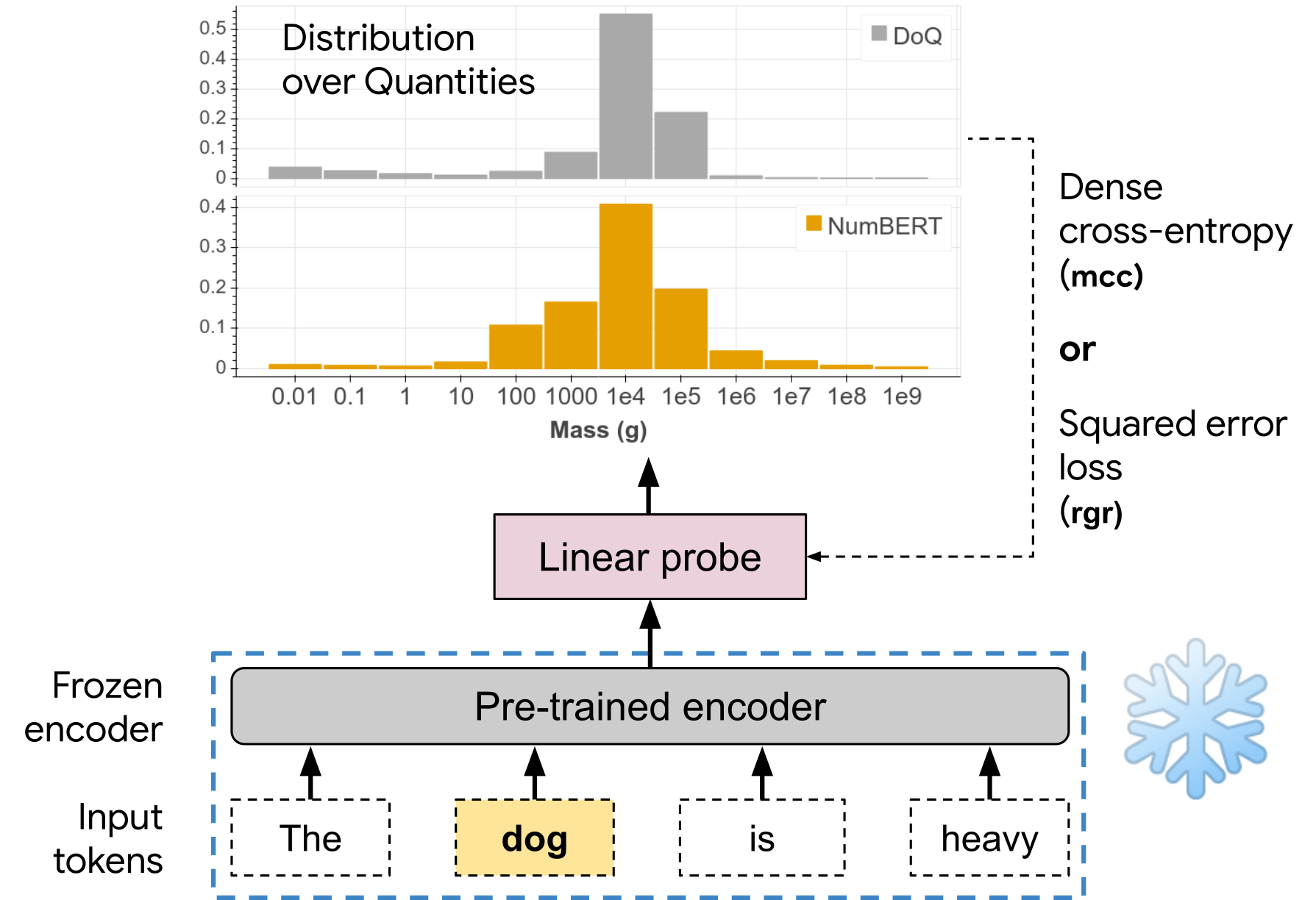}
    \centering
    \caption{Scalar probing example. The mass of ``dog" is a distribution (gray histogram) concentrated around 10-100kg. We train a linear model over a frozen (shown by the snowflake in the figure) encoder to predict this distribution (orange histogram) using either a dense cross-entropy or a regression loss (Section~\ref{sec:probing}).}
    \label{fig:model}
\end{figure}

\begin{figure*}[!ht]
\includegraphics[scale=0.3]{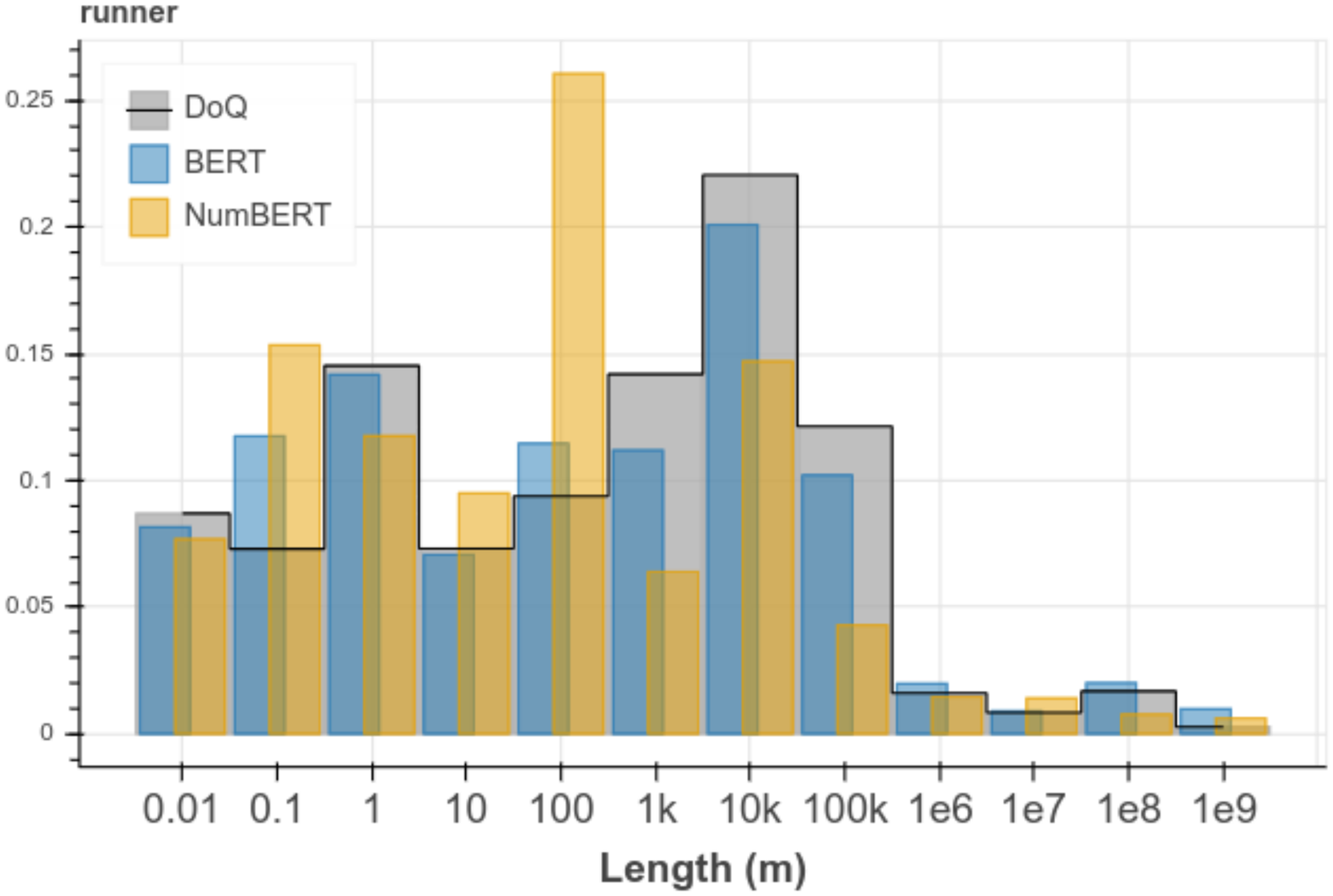}
\includegraphics[scale=0.3]{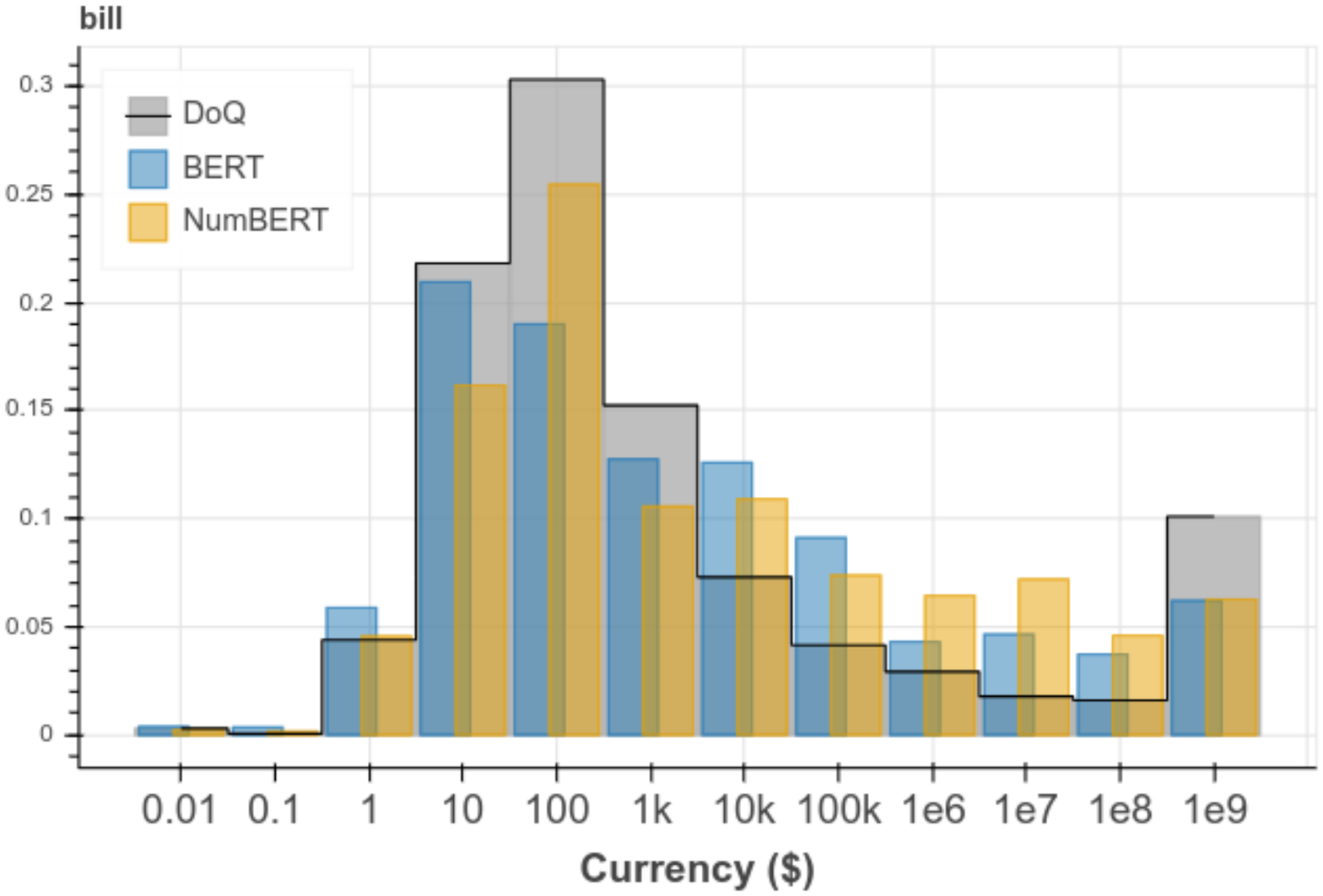}
\caption{Empirical DoQ distributions and scalar probe predictions for MCC+BERT and MCC+NumBERT (Section~\ref{sec:numbert}). The left plot shows length for the term `runner', showing two peaks corresponding to the length of runner cloths and distances run in races. The right plot shows price for the term `bill', with counts corresponding to popular denominations and the volumes of larger currency transactions.}
\label{fig:distributions}
\end{figure*}

Here, we address a simpler task which is another pre-requisite for CSR: the prediction of scalar attributes, a task we call \emph{Scalar Probing}. Given an object (such as a ``wedding ring") and an attribute with continuous numeric values (such as Mass or Price), can an LM's representation of the object predict the value of that attribute? Since in general, there may not be a single correct value for such attributes due to polysemy (``crane'' as a bird, versus construction equipment) or natural variation (e.g. different breeds of dogs), we interpret this as a task of predicting a distribution of possible values for this attribute, and compare it to a ground truth distribution of such values. An overview of this scalar probing is shown in Figure \ref{fig:model}. Examples of ground-truth distributions and model predictions for different objects and attributes are shown in Figure~\ref{fig:distributions}.

 Our analysis shows that contextual encoders, like BERT and ELMo, perform better than noncontextual ones, like Word2Vec, on \textit{scalar probing} despite the task being non-contextual \cite{mikolov2013distributed}. Further, we show that using scientific notation to represent numbers in pre-training can have a significant effect on results (though sensitive to the evaluation metric used). Put together, these results imply that scale representation in contextual encoders is mediated by transfer of magnitude information from numbers to nouns in pre-training and making this mechanism more robust could improve performance on this and other CSR tasks. We also  show improvements by zero-shot transfer from our probes to 2 related tasks: relative comparisons \cite{forbes-choi-2017-verb} and product price prediction \cite{ni2019}, indicating that our results are robust across datasets.


\section{Problem Definition and Data}
We define the scalar probing task  (see Figure \ref{fig:model}) as the problem of predicting a distribution over values of a scalar attribute of an object. We map these values into 12 logarithmically-spaced buckets, so that our task is equivalent to predicting (the distribution of) the order of magnitude of the target value. 
We explore both models that predict the full distribution and models that predict a point estimate of the value, which is essentially a distribution with all the mass concentrating on one bucket.

Our primary resource for the scalar probing task is Distributions over Quantities \citep[DoQ;][]{elazar2019large} which consists of empirical counts of scalar attribute values associated with $>$350K nouns, adjectives, and verbs over 10 different attributes, collected from web data. In this work, we focus only on nouns (which we refer to as \emph{objects}) over the scalar attributes (or \emph{scales}) of MASS (in grams), LENGTH (in meters) and PRICE (in US Dollars). For each object and scale, DoQ provides an empirical distribution over possible values (e.g. Figure~\ref{fig:distributions}) that we map into the 12 afore-mentioned buckets and treat it as ``ground truth".
We note that DoQ itself is derived heuristically from web text and itself contains noise; however, we use it as a starting point to evaluate the performance of different models. Moreover, we validate our findings with transfer experiments shown in Section \ref{sec:discussion}, using DoQ to train a probe that is evaluated on the ground-truth data of \citet{forbes-choi-2017-verb} and \citet{ni2019}. (See Appendix \ref{sec:data_statistics} for detailed  statistics for all 3 datasets.)

To explore the role of context in scalar probing, we also trained  specialized probing models on a subset of DoQ data in narrow domains:  MASS of Animals and PRICE of Household products.

\section{Probing Model}
\label{sec:probing}

We probe three different embedding models: Word2vec \cite{mikolov2013distributed}, ELMo \cite{peters2018deep} and BERT \cite{devlin2018bert} (the latter two of which are contextualized encoders). For each encoder, the input layer extracts an embedding of the object and the probing layer predicts the scalar magnitude. \footnote{We use Word2Vec embeddings of dimension size 500 trained on Wikipedia,  BERT-Base (L=12, H=768, A=12, Total Parameters=110M) trained on Wikipedia+Books and ELMo-Small (LSTM Hidden Size=1024, Output Size=128, \#Highway Layers=1, Total Parameters=13.6M) trained on the 1 Billion Word Benchmark, approximately 800M tokens of news crawl data from WMT 2011.}

\paragraph{Input representations}
For Word2vec, we follow the standard practice of averaging the embeddings of each word in the object's name. If an object name is a full phrase in the  dictionary, we use its embedding instead. As BERT and ELMo are contextual text encoders operating on full sentences, we generate artificial sentences with the following templates:
\begin{itemize}[noitemsep]
    \item \textbf{MASS:} \texttt{The X is heavy.}
    \item \textbf{PRICE:} \texttt{The X is expensive.}
    \item \textbf{LENGTH:} \texttt{The X is big.}
\end{itemize}
and use the CLS token emebedding (for BERT) or final state embedding (for ELMo) as the input representation. For LENGTH, We use ``big" instead of ``long", since LENGTH measurements in DoQ can be widths or heights as well. Variations of these templates with different adjectives and sentence structures (e.g. ``The X is small." or ``What is the length of X?" for LENGTH) led to very similar performance in our evaluations.


\paragraph{Probes}
We use linear probes in all cases following many previous probing work \cite{shi2016does, ettinger2016probing, pimentel2020information} since we want to use a simple probe to find easily accessible information in a representation. \citet{hewitt2019designing} also demonstrates that linear probes achieve relatively high selectivity compared to non-linear ones like MLP.

We experiment with two different approaches for predicting scales:

\subparagraph{Regression (rgr)}
For the point estimate, we use a standard Linear Regression model trained to predict log of the median of all values for each object for the scale attribute under consideration.

\subparagraph{Multi-class Classification (mcc)}

We take a non-parametric approach to modeling the full distribution of scalar values and treat the prediction of which bucket a measurement will fall under as a multi-class classification task, with one class per bucket. A similar approach was shown by \cite{van2016pixel} to perform well for modeling image pixel values. This approach discards the relationship between adjacent bucket values, but it allows us to use the full empirical distribution as soft labels. We train a linear model with softmax output, using a dense cross-entropy loss against the empirical distribution from DoQ.

More details of the models and training procedure are in the Appendix \ref{sec:probing_layer}.


\section{Numeracy through Scientific Notation}
\label{sec:numbert}

\citet{wallace2019numeracy} showed that BERT and ELMo had a limited amount of \emph{numeracy} or numerical reasoning ability, when restricted to numbers of small magnitude. Intuitively, it seems that significant model capacity is expended in parsing the natural representation of numbers as Arabic numerals, where higher and lower order digits are given equal prominence. As further evidence of this, it is shown in Appendix B of \citet{wallace2019numeracy} that the simple intervention of \emph{left-padding} numbers in ELMo instead of the default \emph{right-padding} used in Char-CNNs greatly improves accuracy on these tasks.

To examine the effect of numerical representations on scalar probing, we trained a new version of the BERT model (which we call NumBERT) by replacing every instance of a number in the training data with its representation in \emph{scientific notation}, a combination of an \emph{exponent} and \emph{mantissa} (for example \texttt{314.1} is represented as \texttt{3141[EXP]2} where \texttt{[EXP]} is a new token introduced into the vocabulary). This enables the BERT model to more easily associate objects in the sentence directly with the magnitude expressed in the exponent, ignoring the relatively insignificant mantissa. This model converged to a similar loss on the original BERT Masked LM+NSP pre-training task and a standard suite of NLP tasks (See Appendix \ref{sec:numbert} for more details) as BERT-base, demonstrating that it was not over-specialized for numerical reasoning tasks.

\section{Evaluation} \label{sec:experiments}

We offer the following \textbf{aggregate} baseline to help interpret our results: For each attribute, we compute the empirical distribution over buckets across all objects in the training set, and use that as a predicted distribution for all objects in the test set (this is a stronger version of the majority baseline used in classification tasks).
Since we are comparing results from regression and classification models, we report results on 3 disparate metrics that give a full picture of performance:

\paragraph{Accuracy} For \textbf{mcc} we use the highest scoring bucket from the predicted distribution as the predicted bucket, while for \textbf{rgr} we map the predicted scalar to the single containing bucket and use that as the predicted bucket. Then the accuracy is calculated between the predicted bucket and the ground-truth bucket, which is the highest scoring bucket in the empirical distribution in DoQ.

\paragraph{Mean Square Error (MSE)} When used to compare distributions, this is also known as the \emph{Cramer-Von Mises} distance \cite{baringhaus2017cramer} . It ignores the difference in magnitude between different buckets (a difference in probability mass between buckets $i$ and $i+1$ is equivalent to the same difference between buckets $i$ and any other), but is upper-bounded by 1,  making it easier to interpret. To calculate MSE for \textbf{rgr}, we assume that it assigns a probability of 1 to the single containing bucket.\footnote{This is distinguished from the MSE loss used to train regression models, as it is a distance measure over pairs of distributions.}

\paragraph{Earth Mover's Distance (EMD)} Also known as the \emph{Wasserstein} distance \cite{rubner1998metric}. 




Given two probability densities $p_1$ and $p_2$ on $\Omega$, and some distance measure $\mathrm{d}$ on $\Omega$, the Earth Mover’s Distance is defined as follows:
$$D(p_1, p_2)=\inf_\pi\int_\Omega\int_\Omega \mathrm{d}(x, y) d\pi(x, y)$$
where the infimum is over all non-negative measures $\pi$ on $\Omega\times\Omega$ satisfying $\pi(E\times\Omega)-\pi(\Omega\times E)=\int_E p_1(x)dx-\int_E p_2(x)dx$ for measurable subsets $E\subset\Omega$. Intuitively, EMD measures how much ``work'' needs to be done to move the probability mass of $p_1$ to $p_2$, while MSE measures pointwise what the difference in densities is. So EMD accounts for the distance between buckets, and predictions to neighboring buckets are penalized less than those further away. 

EMD is favored in the statistics literature because of its better convergence properties \cite{rubner1998metric}, and there is evidence that it is more robust to adversarial perturbations of the data distribution \cite{TAT_2019_ICML}, which is relevant for our transfer tasks described below.


\paragraph{Transfer experiments}
\begin{table}
    \small
    \resizebox{\columnwidth}{!}{\begin{tabular}{ll|rr|rr|rr}
    \toprule
               &  & \multicolumn{2}{c}{Accuracy} & \multicolumn{2}{c}{MSE} & \multicolumn{2}{c}{EMD} \\
               &  &      mcc &  rgr &  mcc &  rgr &  mcc &  rgr \\
    \midrule
    \multirow{5}{*}{\rotatebox{90}{\textbf{Lengths}}} & \textbf{Aggregate} &     .24 & .24 & .027 & .027 & .077 & .077 \\
               & \textbf{word2vec} &     .30 & .12 & .026 & .099 & .079 & .072 \\
               & \textbf{ELMo} &     \textbf{.43} & .23 & \textbf{.019} & .084 & .055 & .072 \\
               & \textbf{BERT} & .42 & .24 & .020 & .084 & .056 & .072 \\
               & \textbf{NumBERT} &     .40 & .22 & .021 & .086 & \textbf{.052} & .072 \\
    \cline{1-8}
    \multirow{5}{*}{\rotatebox{90}{\textbf{Masses}}} & \textbf{Aggregate} &     .15 & .15 & .026 & .026 & .076 & .076 \\
               & \textbf{word2vec} &     .26 & .20 & .025 & .088 & .082 & .077 \\
               & \textbf{ELMo} &     \textbf{.36} & .21 & \textbf{.021} & .087 & .061 & .077 \\
               & \textbf{BERT} &     .33 & .22 & \textbf{.021} & .085 & .062 & .077 \\
               & \textbf{NumBERT} &     .32 & .20 & \textbf{.021} & .088 & \textbf{.057} & .077 \\
    \cline{1-8}
    \multirow{5}{*}{\rotatebox{90}{\textbf{Prices}}} & \textbf{Aggregate} &     .24 & .24 & .019 & .019 & .057 & .057 \\
               & \textbf{word2vec} &     .26 & .14 & .019 & .090 & .063 & .087 \\
               & \textbf{ELMo} &     \textbf{.37} & .21 & \textbf{.016} & .081 & .051 & .087 \\
               & \textbf{BERT} &     .33 & .19 & .017 & .083 & .054 & .087 \\
               & \textbf{NumBERT} &     .32 & .17 & .017 & .085 & \textbf{.051} & .087 \\
    \cline{1-8}
    \multirow{5}{*}{\rotatebox{90}{\parbox{1.2cm}{\textbf{Animal Masses}}}} & \textbf{Aggregate} &     .30 & .30 & .022 & .022 & .059 & .059 \\
               & \textbf{word2vec} &     .33 & .35 & .021 & .069 & .069 & .077 \\
               & \textbf{ELMo} &     \textbf{.43} & .28 & \textbf{.016} & .079 & .057 & .077 \\
               & \textbf{BERT} &     .41 & .26 & .017 & .079 & .058 & .077 \\
               & \textbf{NumBERT} &     .42 & .23 & .018 & .083 & \textbf{.053} & .077 \\
    \bottomrule
    \end{tabular}}
    \caption{Comparison of encoders and probes on the Scalar probing task on DoQ data. Results are averaged over 10-fold cross-validation.}
    \label{tab:main_results}
\end{table}
We also evaluate models trained on DoQ on 2 datasets containing ground truth labels of scalar attributes. The first is a human-labeled dataset of \textit{relative comparisons} (e.g. \textit{(person, fox, weight, bigger)}) \cite{forbes-choi-2017-verb}. Predictions for this task are made by comparing the point estimates for \textbf{rgr} and highest-scoring buckets for \textbf{mcc}. 
The second is an empirical distribution of product price data extracted from the Amazon Review Dataset \cite{ni2019}. We retrained a model on DoQ prices using 12 power-of-4 buckets to support finer grained predictions.

\section{Results} \label{sec:discussion}

    Table \ref{tab:main_results} shows results of scalar probing on DoQ data.\footnote{The full set of experimental results are shown in Table \ref{tab:big_results} in Appendix \ref{sec:complete_results}.}
    For \textbf{MSE} and \textbf{EMD} the best possible score is 0, while for accuracy we take a loose upper bound to be the performance of a model that samples from the ground-truth distribution and is evaluated against the mode. This method achieves accuracies of 0.570 for lengths, 0.537 for masses, and 0.476 for prices. Compared to the baseline, we can see that \textbf{mcc} over the best encoders capture about half (as measured by accuracy) to a third (by \textbf{MSE} and \textbf{EMD}) of the distance to the upper bound, suggesting that while a significant amount of scalar information is available, there is a long way to go to support robust commonsense reasoning.
    
    From Table \ref{tab:main_results}, we see that the more expressive models using \textbf{mcc} consistently beat \textbf{rgr}, with the latter frequently unable to improve upon the Aggregate baseline.  This shows that scale information is present in the embeddings, but training on the median alone is not enough to reliably extract it; the full data distribution is needed.
    
    Comparing results by encoders, we see that Word2Vec performs significantly worse than the contextual encoders -- even though the task is non-contextual -- indicating that contextual information during pre-training improves the representation of scales.

    Despite being weaker than BERT on downstream NLP tasks, ELMo does better on scalar probing, consistent with it being better at numeracy \cite{wallace2019numeracy} due to its character-level tokenization. 

    NumBERT does consistently better than ELMo and BERT on the \textbf{EMD} metric, but worse on \textbf{MSE} and Accuracy. This is in contrast to other standard benchmarks such as Q/A and NLI, where NumBERT made no difference relative to BERT. 
    Our key takeaway is that the numerical representation has an impact on scale prediction (see Figure~\ref{fig:distributions} for qualitative differences), but the direction is sensitive to the choice of evaluation metric. As discussed in Section \ref{sec:experiments}, we believe EMD to be the most robust metric \textit{a priori}, but this finding highlights the need to still examine the full range of metrics.

Results on Animal Masses (Table \ref{tab:main_results}) show that training models only on objects in a narrow domain can significantly improve scalar prediction, underscoring the importance of context. For example, while ``crane'' in general can refer to either a bird or a piece of construction equipment, only the former is relevant in the animal domain, giving the model a simpler distribution of masses to predict.
    
    Note that, despite significant differences in the raw numbers for each scale (mass/length/price), the relative behavior of encoders, metrics and probes are the same, indicating that our conclusions are broadly applicable.

\paragraph{Transfer experiments}
      On the F\&C  relative comparison task (Table \ref{tab:verb_physics}),  \textbf{rgr}+NumBERT performed best, approaching the performance of using DoQ as an oracle, though short of specialized models for this task \cite{yang-etal-2018-extracting}. Scalar probes trained with \textbf{mcc} perform poorly, possibly because a finer-grained model of predicted distribution is not useful for the 3-class comparative task. On the Amazon price dataset (Table \ref{tab:product_prices}) which is a full distribution prediction task, \textbf{mcc}+NumBERT did best on both distributional metrics. On both zero-shot transfer tasks, NumBERT was the best encoder on all configurations of metric/objective, suggesting that manipulating numeric representations can significantly improve performance on scalar prediction.


\begin{table}
    \centering
    \begin{tabular}{lrrrr}
    \toprule
                         & \multicolumn{2}{c}{dev} & \multicolumn{2}{c}{test} \\
                          & mcc & rgr &  mcc & rgr \\
    \midrule
    \textbf{word2vec    } & .40 & .73 &  .38 & .74 \\
    \textbf{ELMo        } & .47 & .71 &  .47 & .72 \\
    \textbf{BERT        } & .48 & .71 &  .49 & .71 \\
    \textbf{NumBERT     } & .51 & .77 &  .54 & .76 \\
    \textbf{\small DoQ [Elazar et. al. 2019]     } & - & .78 & -  & .77 \\
    \textbf{Yang et. al. '18} & - & \textbf{.86} & -  & \textbf{.87} \\
    \bottomrule
    \end{tabular}
    \caption{Accuracy on VerbPhysics \cite{forbes-choi-2017-verb}.
    }
    \label{tab:verb_physics}
\end{table}

\begin{table}
    \begin{adjustwidth}{-.25in}{-.25in}
    \centering
    \begin{tabular}{lrrrrrr}
    \toprule
    {} & \multicolumn{2}{c}{Accuracy} & \multicolumn{2}{c}{MSE} & \multicolumn{2}{c}{EMD} \\
                           &      mcc & rgr & mcc & rgr & mcc & rgr \\
    \midrule
    \textbf{Aggregate   } &      .04 & .04 & \textbf{.02} & \textbf{.02} & .06 & .06 \\
    \textbf{word2vec    } &      .09 & .23 & \textbf{.02} & .07 & .07 & .08 \\
    \textbf{BERT        } &      .14 & .25 & \textbf{.02} & .07 & .06 & .08 \\
    \textbf{NumBERT     } &      .18 & \textbf{.27} & \textbf{.02} & .07 & \textbf{.05} & .08 \\
    \bottomrule
    \end{tabular}
    \end{adjustwidth}
    \caption{Results on consumer price data \cite{ni2019}.}
    \label{tab:product_prices}
\end{table}

\section{Conclusion}

 From our novel scalar probing experiments, we find there is a significant amount of scale information in object embeddings, but still a sizable gap to overcome before LMs achieve this prerequisite of CSR. 
 We conclude that although we observe some non-trivial signal to extract scale information from language embedding, the weak signals suggest these models are far from satisfying common sense scale understanding.
 
 Our analysis points to improvements in modeling context and numeracy as directions in which progress can be made, mediated by the transfer of scale information from numbers to nouns. 
 The NumBERT intervention has a measurable impact on scalar probing results, and transfer experiments suggest that it is an improvement. For future work we would like to extend our models to predict scales for sentences bearing relevant context about scalar magnitudes, e.g. ``I saw a baby elephant".

\section*{Acknowledgments}
We want to thank Daniel Spokoyny and William Cohen for the idea of using scientific notation for numbers and Jeremiah Liu for helpful discussions on statistical distance measures.

\bibliographystyle{acl_natbib}
\bibliography{anthology,acl2020}

\begin{thebibliography}{20}
\expandafter\ifx\csname natexlab\endcsname\relax\def\natexlab#1{#1}\fi

\bibitem[{Baringhaus and Henze(2017)}]{baringhaus2017cramer}
L~Baringhaus and N~Henze. 2017.
\newblock Cram{\'e}r--von mises distance: probabilistic interpretation,
  confidence intervals, and neighbourhood-of-model validation.
\newblock \emph{Journal of Nonparametric Statistics}, 29(2):167--188.

\bibitem[{Bisk et~al.(2020)Bisk, Zellers, Bras, Gao, and Choi}]{bisk2020}
Yonatan Bisk, Rowan Zellers, Ronan~Le Bras, Jianfeng Gao, and Yejin Choi. 2020.
\newblock Piqa: Reasoning about physical commonsense in natural language.
\newblock In \emph{Thirty-Fourth AAAI Conference on Artificial Intelligence}.

\bibitem[{Devlin et~al.(2018)Devlin, Chang, Lee, and
  Toutanova}]{devlin2018bert}
Jacob Devlin, Ming-Wei Chang, Kenton Lee, and Kristina Toutanova. 2018.
\newblock Bert: Pre-training of deep bidirectional transformers for language
  understanding.
\newblock \emph{arXiv preprint arXiv:1810.04805}.

\bibitem[{Elazar et~al.(2019)Elazar, Mahabal, Ramachandran, Bedrax-Weiss, and
  Roth}]{elazar2019large}
Yanai Elazar, Abhijit Mahabal, Deepak Ramachandran, Tania Bedrax-Weiss, and Dan
  Roth. 2019.
\newblock How large are lions? inducing distributions over quantitative
  attributes.
\newblock In \emph{Association for Computational Linguistics (ACL)}.

\bibitem[{Ettinger et~al.(2016)Ettinger, Elgohary, and
  Resnik}]{ettinger2016probing}
Allyson Ettinger, Ahmed Elgohary, and Philip Resnik. 2016.
\newblock Probing for semantic evidence of composition by means of simple
  classification tasks.
\newblock In \emph{Proceedings of the 1st Workshop on Evaluating Vector-Space
  Representations for NLP}, pages 134--139.

\bibitem[{Forbes and Choi(2017)}]{forbes-choi-2017-verb}
Maxwell Forbes and Yejin Choi. 2017.
\newblock \href {https://doi.org/10.18653/v1/P17-1025} {Verb physics: Relative
  physical knowledge of actions and objects}.
\newblock In \emph{Proceedings of the 55th Annual Meeting of the Association
  for Computational Linguistics (Volume 1: Long Papers)}, pages 266--276,
  Vancouver, Canada. Association for Computational Linguistics.

\bibitem[{Geva et~al.(2020)Geva, Gupta, and Berant}]{geva2020injecting}
Mor Geva, Ankit Gupta, and Jonathan Berant. 2020.
\newblock Injecting numerical reasoning skills into language models.
\newblock \emph{arXiv preprint arXiv:2004.04487}.

\bibitem[{Hewitt and Liang(2019)}]{hewitt2019designing}
John Hewitt and Percy Liang. 2019.
\newblock Designing and interpreting probes with control tasks.
\newblock \emph{arXiv preprint arXiv:1909.03368}.

\bibitem[{Jianmo~Ni(2019)}]{ni2019}
Julian~McAuley Jianmo~Ni, Jiacheng~Li. 2019.
\newblock Justifying recommendations using distantly-labeled reviews and
  fined-grained aspects.
\newblock In \emph{Empirical Methods in Natural Language Processing (EMNLP)}.

\bibitem[{Lin et~al.(2020)Lin, Lee, Khanna, and Ren}]{lin2020birds}
Bill~Yuchen Lin, Seyeon Lee, Rahul Khanna, and Xiang Ren. 2020.
\newblock Birds have four legs?! numersense: Probing numerical commonsense
  knowledge of pre-trained language models.
\newblock \emph{arXiv preprint arXiv:2005.00683}.

\bibitem[{Liu et~al.(2019)Liu, Long, Wang, and Jordan}]{TAT_2019_ICML}
Hong Liu, Mingsheng Long, Jianmin Wang, and Michael~I. Jordan. 2019.
\newblock Transferable adversarial training: A general approach to adapting
  deep classifiers.
\newblock In \emph{Proceedings of the 36th International Conference on Machine
  Learning}.

\bibitem[{Mikolov et~al.(2013)Mikolov, Sutskever, Chen, Corrado, and
  Dean}]{mikolov2013distributed}
Tomas Mikolov, Ilya Sutskever, Kai Chen, Greg~S Corrado, and Jeff Dean. 2013.
\newblock Distributed representations of words and phrases and their
  compositionality.
\newblock In \emph{Advances in neural information processing systems}, pages
  3111--3119.

\bibitem[{Peters et~al.(2018)Peters, Neumann, Iyyer, Gardner, Clark, Lee, and
  Zettlemoyer}]{peters2018deep}
Matthew~E Peters, Mark Neumann, Mohit Iyyer, Matt Gardner, Christopher Clark,
  Kenton Lee, and Luke Zettlemoyer. 2018.
\newblock Deep contextualized word representations.
\newblock \emph{arXiv preprint arXiv:1802.05365}.

\bibitem[{Pimentel et~al.(2020)Pimentel, Valvoda, Maudslay, Zmigrod, Williams,
  and Cotterell}]{pimentel2020information}
Tiago Pimentel, Josef Valvoda, Rowan~Hall Maudslay, Ran Zmigrod, Adina
  Williams, and Ryan Cotterell. 2020.
\newblock Information-theoretic probing for linguistic structure.
\newblock \emph{arXiv preprint arXiv:2004.03061}.

\bibitem[{Rubner et~al.(1998)Rubner, Tomasi, and Guibas}]{rubner1998metric}
Yossi Rubner, Carlo Tomasi, and Leonidas~J Guibas. 1998.
\newblock A metric for distributions with applications to image databases.
\newblock In \emph{Sixth International Conference on Computer Vision (IEEE Cat.
  No. 98CH36271)}, pages 59--66. IEEE.

\bibitem[{Shi et~al.(2016)Shi, Padhi, and Knight}]{shi2016does}
Xing Shi, Inkit Padhi, and Kevin Knight. 2016.
\newblock Does string-based neural mt learn source syntax?
\newblock In \emph{Proceedings of the 2016 Conference on Empirical Methods in
  Natural Language Processing}, pages 1526--1534.

\bibitem[{Van~Oord et~al.(2016)Van~Oord, Kalchbrenner, and
  Kavukcuoglu}]{van2016pixel}
Aaron Van~Oord, Nal Kalchbrenner, and Koray Kavukcuoglu. 2016.
\newblock Pixel recurrent neural networks.
\newblock In \emph{International Conference on Machine Learning}, pages
  1747--1756.

\bibitem[{Wallace et~al.(2019)Wallace, Wang, Li, Singh, and
  Gardner}]{wallace2019numeracy}
Eric Wallace, Yizhong Wang, Sujian Li, Sameer Singh, and Matt Gardner. 2019.
\newblock \href {https://doi.org/10.18653/v1/D19-1534} {Do {NLP} models know
  numbers? probing numeracy in embeddings}.
\newblock In \emph{Proceedings of the 2019 Conference on Empirical Methods in
  Natural Language Processing and the 9th International Joint Conference on
  Natural Language Processing (EMNLP-IJCNLP)}, pages 5306--5314, Hong Kong,
  China. Association for Computational Linguistics.

\bibitem[{Yang et~al.(2018)Yang, Birnbaum, Wang, and
  Downey}]{yang-etal-2018-extracting}
Yiben Yang, Larry Birnbaum, Ji-Ping Wang, and Doug Downey. 2018.
\newblock \href {https://doi.org/10.18653/v1/P18-2102} {Extracting commonsense
  properties from embeddings with limited human guidance}.
\newblock In \emph{Proceedings of the 56th Annual Meeting of the Association
  for Computational Linguistics (Volume 2: Short Papers)}, pages 644--649,
  Melbourne, Australia. Association for Computational Linguistics.

\bibitem[{Zhou et~al.(2020)Zhou, Ning, Khashabi, and Roth}]{zhao2020}
Ben Zhou, Qiang Ning, Daniel Khashabi, and Dan Roth. 2020.
\newblock Temporal common sense acquisition with minimal supervision.
\newblock In \emph{Association for Computational Linguistics}.

\end{thebibliography}

\appendix



\section{Model Hyperparameters}
Here we provide the model hyperparameters we use for reproducibility.

\subsection{Probing Layer of the Scalar Probing Model} \label{sec:probing_layer}

For \textbf{rgr}, we use a ridge regression with regularization strength of 1. For \textbf{mcc}, we use a linear classifier with a softmax activation function and regularization strength of 0.01.

For experiments on the narrow domains with smaller datasets, we first use PCA to reduce embeddings down to 150 dimensions before training the probing model. 

\subsection{NumBERT} \label{sec:numbert}
NumBERT is pretrained on the Wikipedia and Books corpora used by the original BERT paper \cite{devlin2018bert}. The BERT configuration is the same as BERT-Base (L=12, H=768, A=12, Total Parameters=110M). The language model masking is applied after WordPiece tokenization with a uniform masking rate of 15\%. Maximum sequence length (number of tokens) is 128. We train with batch size of 64 sequences for 1,000,000 steps, which is approximately 40 epochs over the 3.3 billion word corpus. All the other hyperparameters and implementation details (optimizer, warm-up steps, etc.) are the same as the original BERT implementation. Table \ref{tab:numbertvbert} shows a comparison of NumBERT vs a re-implementation of BERT-base with identical settings as above, on a suite of standard NLP benchmarks, and we conclude that the two models reach similar performance on these tasks.

\section{Data Statistics} \label{sec:data_statistics}

Table \ref{tab:data-statistics} shows the statistics of 3 datasets/resources we use in this paper. For DoQ, we take the original resource and get each subset by filtering using the corresponding dimensions and/or object types (e.g. all objects, animals, product categories, etc.). Also, only objects with more than 100 values collected in the resource are used. For F\&C Cleaned dataset, we use the data and the train/dev/test splits from \cite{elazar2019large}.

\begin{table}
    \begin{adjustwidth}{-.25in}{-.25in}
    \setlength{\tabcolsep}{2pt} 
    \centering
    \begin{tabular}{|l|l|c|c|}
    \hline
    Task & Metric & BERT  & NumBERT \\
    & & Base & \\
    \hline
    CoLA     & Dev Acc & .745 &  .742  \\
    MNLI         & {\small Dev Matched Acc} & .791 &  .789 \\
     & {\small Dev Mismatched Acc} & .795 &  .798 \\
    MRPC         & Dev Acc & .816 &  .802 \\
    Squad v1      & F1 & .799 &  .789\\
    Squad v2     & Best F1 & .669 & .673 \\
    STS-B & Dev Pearson's r & .866 & .871 \\
    \hline
    \end{tabular}
    \end{adjustwidth}
    \caption{NumBERT vs BERT-base on a suite of standard NLP benchmarks.}
    \label{tab:numbertvbert}
\end{table}

\begin{table}[]
    \setlength{\tabcolsep}{2pt} 
    \centering
    \resizebox{\columnwidth}{!}{\begin{tabular}{|l|l|r|}
    \hline
    Dataset & Subset & \#Data Samples \\ \hline
    \multirow{5}{*}{DoQ} & all masses & 76,424 \\ \cline{2-3} 
     & all prices & 212,277 \\ \cline{2-3} 
     & all lengths & 244,517 \\ \cline{2-3} 
     & animal masses & 519 \\ \cline{2-3} 
     & product category prices & 1,789 \\ \hline
    Product Price & - & 1,888 \\ \hline
    \multirow{3}{*}{F\&C Cleaned} & train & 172 \\ \cline{2-3} 
     & dev & 1,267 \\ \cline{2-3} 
     & test & 1,522 \\ \hline
    \end{tabular}}
    \caption{Statistics of Datasets/Resources used in our paper}
    \label{tab:data-statistics}
\end{table}

\section{Complete Experiment Results} \label{sec:complete_results}

We model the distributions of those scalar attributes as categorical distributions over 12 categories. We first take the base-10 logarithm of all the values and then round them to the nearest integer (between -2 and 9 for all scales). We treat each integer as a bucket and use the normalized counts in each bucket as the true distribution for that scalar attribute of the object. 

To explore the effect of ambiguity, we divide all the data in DoQ for each scale into 2 sets, \textbf{Unimodal} where the distribution has one well-defined peak and \textbf{Multimodal}, where multiple peaks are present. The number of peaks were identified by a simple hill-climbing algorithm.

As words often have more than one meaning in different contexts  or even modifiers, their corresponding distribution from DoQ should reflect the different senses if they appeared enough in the data. When the objects are different enough (e.g. an ice-cream have mainly one meaning and its size doesn't vary much, as opposed to a truck which can be a toy truck, which is very small, or an actual vehicle, which is very big), they may have different modalities. In order to better understand our results, we wish to separate between objects with multiple modalities from those with a single modality.

In order to estimate a multi-modal function, we take the bucketed DoQ distribution and smooth it into a probability density function. Then, by finding local maxima over the fitted density function, we estimate a distribution to be multi-modal if we find more than one maximum.

The complete experiment results including the mutlimodal experiments are in Table \ref{tab:big_results}.

\begin{table*}
\centering
\begin{tabular}{lllrrrrrrrrr}
\toprule
                         &     &  & \multicolumn{3}{c}{\textbf{Accuracy}} & \multicolumn{3}{c}{\textbf{MSE}} & \multicolumn{3}{c}{\textbf{EMD}} \\
                         &     &  &      All & Multi. & Uni. & All & Multi. & Uni. & All & Multi. & Uni. \\
\midrule

\multirow{10}{*}{\rotatebox{90}{\textbf{Lengths}}} & \multirow{5}{*}{\textbf{mcc}} & \textbf{Aggregate} &     .240 &       .250 &     .230 & .027 &       .028 &     .025 & .077 &       .078 &     .075 \\
                         &     & \textbf{word2vec} &     .300 &       .310 &     .280 & .026 &       .022 &     .031 & .079 &       .074 &     .087 \\
                         &     & \textbf{ELMo} &     .430 &       .420 &     .440 & .019 &       .020 &     .017 & .055 &       .056 &     .053 \\
                         &     & \textbf{BERT} &     .420 &       .410 &     .420 & .020 &       .021 &     .018 & .056 &       .058 &     .054 \\
                         &     & \textbf{NumBERT} &     .400 &       .400 &     .410 & .021 &       .022 &     .019 & .052 &       .053 &     .049 \\
\cline{2-12}
                         & \multirow{5}{*}{\textbf{rgr}} & \textbf{Aggregate} &     .240 &       .250 &     .230 & .027 &       .028 &     .025 & .077 &       .078 &     .075 \\
                         &     & \textbf{word2vec} &     .120 &       .120 &     .130 & .099 &       .100 &     .097 & .072 &       .070 &     .074 \\
                         &     & \textbf{ELMo} &     .230 &       .230 &     .240 & .084 &       .085 &     .082 & .072 &       .070 &     .074 \\
                         &     & \textbf{BERT} &     .240 &       .230 &     .240 & .084 &       .085 &     .081 & .072 &       .070 &     .074 \\
                         &     & \textbf{NumBERT} &     .220 &       .210 &     .220 & .086 &       .088 &     .084 & .072 &       .070 &     .074 \\
\cline{1-12}
\cline{2-12}
\multirow{10}{*}{\rotatebox{90}{\textbf{Masses}}} & \multirow{5}{*}{\textbf{mcc}} & \textbf{Aggregate} &     .150 &       .150 &     .150 & .026 &       .027 &     .024 & .076 &       .077 &     .074 \\
                         &     & \textbf{word2vec} &     .260 &       .260 &     .260 & .025 &       .026 &     .023 & .082 &       .083 &     .080 \\
                         &     & \textbf{ELMo} &     .360 &       .360 &     .360 & .021 &       .021 &     .019 & .061 &       .062 &     .059 \\
                         &     & \textbf{BERT} &     .330 &       .330 &     .330 & .021 &       .022 &     .019 & .062 &       .063 &     .060 \\
                         &     & \textbf{NumBERT} &     .320 &       .320 &     .330 & .021 &       .022 &     .019 & .057 &       .058 &     .055 \\
\cline{2-12}
                         & \multirow{5}{*}{\textbf{rgr}} & \textbf{Aggregate} &     .150 &       .150 &     .150 & .026 &       .027 &     .024 & .076 &       .077 &     .074 \\
                         &     & \textbf{word2vec} &     .200 &       .190 &     .200 & .088 &       .090 &     .086 & .077 &       .076 &     .080 \\
                         &     & \textbf{ELMo} &     .210 &       .200 &     .210 & .087 &       .088 &     .085 & .077 &       .076 &     .080 \\
                         &     & \textbf{BERT} &     .220 &       .210 &     .220 & .085 &       .086 &     .084 & .077 &       .076 &     .080 \\
                         &     & \textbf{NumBERT} &     .200 &       .190 &     .200 & .088 &       .089 &     .086 & .077 &       .076 &     .080 \\
\cline{1-12}
\cline{2-12}
\multirow{10}{*}{\rotatebox{90}{\textbf{Prices}}} & \multirow{5}{*}{\textbf{mcc}} & \textbf{Aggregate} &     .240 &       .240 &     .250 & .019 &       .021 &     .016 & .057 &       .060 &     .054 \\
                         &     & \textbf{word2vec} &     .260 &       .250 &     .280 & .019 &       .014 &     .024 & .063 &       .055 &     .072 \\
                         &     & \textbf{ELMo} &     .370 &       .360 &     .380 & .016 &       .018 &     .013 & .051 &       .053 &     .047 \\
                         &     & \textbf{BERT} &     .330 &       .320 &     .330 & .017 &       .019 &     .014 & .054 &       .055 &     .051 \\
                         &     & \textbf{NumBERT} &     .320 &       .320 &     .330 & .017 &       .019 &     .014 & .051 &       .053 &     .048 \\
\cline{2-12}
                         & \multirow{5}{*}{\textbf{rgr}} & \textbf{Aggregate} &     .240 &       .240 &     .250 & .019 &       .021 &     .016 & .057 &       .060 &     .054 \\
                         &     & \textbf{word2vec} &     .140 &       .130 &     .150 & .090 &       .093 &     .085 & .087 &       .084 &     .092 \\
                         &     & \textbf{ELMo} &     .210 &       .210 &     .220 & .081 &       .083 &     .078 & .087 &       .084 &     .092 \\
                         &     & \textbf{BERT} &     .190 &       .190 &     .190 & .083 &       .085 &     .081 & .087 &       .084 &     .092 \\
                         &     & \textbf{NumBERT} &     .170 &       .180 &     .170 & .085 &       .087 &     .083 & .087 &       .084 &     .092 \\
\cline{1-12}
\cline{2-12}
\multirow{10}{*}{\rotatebox{90}{\textbf{Animals Masses}}} & \multirow{5}{*}{\textbf{mcc}} & \textbf{Aggregate} &     .300 &       .280 &     .330 & .022 &       .021 &     .024 & .059 &       .055 &     .064 \\
                         &     & \textbf{word2vec} &     .330 &       .320 &     .350 & .021 &       .020 &     .023 & .069 &       .066 &     .075 \\
                         &     & \textbf{ELMo} &     .430 &       .440 &     .420 & .016 &       .015 &     .019 & .057 &       .056 &     .059 \\
                         &     & \textbf{BERT} &     .410 &       .390 &     .450 & .017 &       .016 &     .019 & .058 &       .057 &     .060 \\
                         &     & \textbf{NumBERT} &     .420 &       .430 &     .410 & .018 &       .016 &     .020 & .053 &       .052 &     .055 \\
\cline{2-12}
                         & \multirow{5}{*}{\textbf{rgr}} & \textbf{Aggregate} &     .300 &       .280 &     .330 & .022 &       .021 &     .024 & .059 &       .055 &     .064 \\
                         &     & \textbf{word2vec} &     .350 &       .350 &     .360 & .069 &       .069 &     .069 & .077 &       .081 &     .070 \\
                         &     & \textbf{ELMo} &     .280 &       .250 &     .330 & .079 &       .080 &     .077 & .077 &       .081 &     .070 \\
                         &     & \textbf{BERT} &     .260 &       .260 &     .240 & .079 &       .076 &     .085 & .077 &       .081 &     .070 \\
                         &     & \textbf{NumBERT} &     .230 &       .230 &     .240 & .083 &       .081 &     .086 & .077 &       .081 &     .070 \\
\cline{1-12}
\cline{2-12}
\multirow{10}{*}{\rotatebox{90}{\textbf{Household Product Prices}}} & \multirow{5}{*}{\textbf{mcc}} & \textbf{Aggregate} &     .470 &        - &      - & .010 &        - &      - & .046 &        - &      - \\
                         &     & \textbf{word2vec} &     .510 &       .490 &     .540 & .008 &       .008 &     .008 & .041 &       .041 &     .041 \\
                         &     & \textbf{ELMo} &     .540 &       .520 &     .570 & .007 &       .007 &     .007 & .038 &       .038 &     .039 \\
                         &     & \textbf{BERT} &     .570 &       .560 &     .580 & .007 &       .007 &     .007 & .038 &       .038 &     .039 \\
                         &     & \textbf{NumBERT} &     .550 &       .530 &     .570 & .007 &       .007 &     .007 & .038 &       .038 &     .039 \\
\cline{2-12}
                         & \multirow{5}{*}{\textbf{rgr}} & \textbf{Aggregate} &     .470 &        - &      - & .010 &        - &      - & .046 &        - &      - \\
                         &     & \textbf{word2vec} &     .450 &       .430 &     .480 & .056 &       .058 &     .055 & .092 &       .094 &     .090 \\
                         &     & \textbf{ELMo} &     .420 &       .400 &     .460 & .058 &       .059 &     .057 & .092 &       .094 &     .090 \\
                         &     & \textbf{BERT} &     .440 &       .420 &     .460 & .057 &       .059 &     .055 & .092 &       .094 &     .090 \\
                         &     & \textbf{NumBERT} &     .420 &       .390 &     .460 & .060 &       .062 &     .057 & .092 &       .094 &     .090 \\
\bottomrule
\end{tabular}
\caption{Evaluation on all datasets.}
\label{tab:big_results}
\end{table*}


\end{document}